\documentclass[runningheads]{llncs}

\usepackage{graphicx}
\usepackage{amsmath,amssymb} 

\usepackage{color}
\usepackage{xcolor}
\usepackage{tabularx}
\usepackage{subfig}
\usepackage{caption}

\usepackage{array}
\usepackage{latexsym}

\makeatletter
\def\blfootnote{\gdef\@thefnmark{}\@footnotetext}
\makeatother

\newcolumntype{L}[1]{>{\raggedright\let\newline\\\arraybackslash\hspace{0pt}}m{#1}}
\newcolumntype{C}[1]{>{\centering\let\newline\\\arraybackslash\hspace{0pt}}m{#1}}

\definecolor{orange}{rgb}{1,0.5,0}
\definecolor{deeppink}{RGB}{255,20,147}

\newif\ifdraft

\draftfalse

\ifdraft

\else

\fi

\newcommand{\pposes}{\mathbf{X}_{-M:-1}}
\newcommand{\pposesA}{\mathbf{X}_{-M_1:-1}}
\newcommand{\pposesB}{\mathbf{X}_{-M_2:-1}}
\newcommand{\pposesj}{\mathbf{X}_{-M_j:-1}}

\newcommand{\fposes}{\mathbf{X}_{0:T-1}}

\newcommand{\pose}{\mathbf{X}}
\newcommand{\embed}{\mathbf{E}}

\newcommand{\R}{\mathbb{R}}

\begin{document}
\pagestyle{headings}
\mainmatter

\def\ACCV20SubNumber{857}

\author{Tim Lebailly\inst{1}\orcidID{0000-0002-9814-7531} \and
Sena Kiciroglu\inst{1}\orcidID{0000-0003-2739-804X} \and
Mathieu Salzmann\inst{1,2}\orcidID{0000-0002-8347-8637} \and
Pascal Fua\inst{1}\orcidID{0000-0002-5477-1017} \and
\\Wei Wang\inst{1,3}\orcidID{0000-0002-5477-1017}\thanks{Corresponding author: Wei Wang} }

\authorrunning{T. Lebailly et al.}

\institute{CVLab EPFL, Switzerland \\
	\email{\{firstname.lastname\}@epfl.ch}\and
ClearSpace, Switzerland \and
University of Trento, Italy}

\title{Motion Prediction Using Temporal Inception Module} 
\titlerunning{Motion Prediction Using TIM}

\maketitle

\begin{abstract}
Human motion prediction is a necessary component for many applications in robotics and autonomous driving. Recent methods propose using sequence-to-sequence deep learning models to tackle this problem. However, they do not focus on exploiting different temporal scales for different length inputs. We argue that the diverse temporal scales are important as they allow us to look at the past frames with different receptive fields, which can lead to better predictions. In this paper, we propose a Temporal Inception Module (TIM) to encode human motion. Making use of TIM, our framework produces input embeddings using convolutional layers, by using different kernel sizes for different input lengths. The experimental results on standard motion prediction benchmark datasets Human3.6M and CMU motion capture dataset show that our approach consistently outperforms the state of the art methods.
\end{abstract}

\section{Introduction}

Human motion prediction is an essential component for a wide variety of applications. For instance, in the field of robotics, robots working closely with humans require an internal representation of the current and future human motion to navigate around them safely \cite{Gui18}. Autonomous driving is another important use case where cars need to forecast pedestrian motion accurately to avoid accidents \cite{Habibi2018ContextAwarePM,Fan13}. Other applications such as sports tracking also heavily use these forecasting methods for better performances \cite{Kiciroglu20}.

In order to achieve high accuracy motion prediction, we show that the encoding of the body joint trajectories (i.e., sequence of 3D joint locations) is key. In \cite{Mao19} this is achieved by representing each trajectory using its Discrete Cosine Transform (DCT) coefficients \cite{1672377}, a technique previously used to encode human motion for human pose estimation \cite{Lin19,Huang18d}. However, we show that we can gain a large boost in accuracy by using a network to encode the trajectories at multiple temporal scales. In particular, inspired by the Inception Module of \cite{Szegedy15}, we have created a ``Temporal Inception Module'', which uses various size convolutional kernels to filter the trajectory at different temporal scales for different input sizes. This allows the network have different receptive fields in the temporal domain.

Following \cite{Li20,Mao19}, the backbone of our prediction architecture is based on a graph convolutional network (GCN) \cite{Bruna14} which is a high capacity feed-forward model. As input to the GCN, Mao \emph{et al.} \cite{Mao19} transform time sequences of joint locations from the $10$ past frames into a DCT representation. Moreover, they demonstrate that more frames from past do not help to boost the performance. In our paper we show that by looking at the trajectory at a multiple temporal scale, more frames from the past actually do help to further improve the performance, which is especially true for long-term future motion prediction. Therefore, instead of using the DCT coefficients of the trajectory as the input to the GCN, we use an encoder module to produce the input embeddings at multiple temporal scales. 

Our key idea lies in the fact that recently seen frames hold more relevant information for the prediction of the near future frames than older ones that are far away from the current frame. Therefore by having many smaller kernels that look specifically at recent frames we are able to place more emphasis on the recent frames. This is especially useful for short-term prediction. Nevertheless, for long-term future frame prediction, the older frames also become important as they are able to describe the high-level motion patterns. For instance, for a walking motion which contains the pattern of moving left and right legs in turn, the most recently seen frames only contain the motion of one leg, rather than the cyclic motion of both legs. These high-level motion patterns are usually lower frequency signals. Incorporating this prior knowledge in the encoding of the trajectory allows us to keep local features of the recently seen frames while also keeping the high-level motion pattern for older frames. This inductive bias gives us a boost in accuracy.

In summary, our contributions are twofold: 
\begin{itemize}
    \item We introduce the Temporal Inception Module (TIM), which allows the network to view the motion trajectory at different temporal scales which leads to better performance.
    \item We present our action-agnostic end to end trainable pipeline combining TIM and GCN which can be trained once to handle all actions evaluated. 
\end{itemize}

\begin{figure}[h!]
\centering
 
\includegraphics[width=0.885\textwidth]{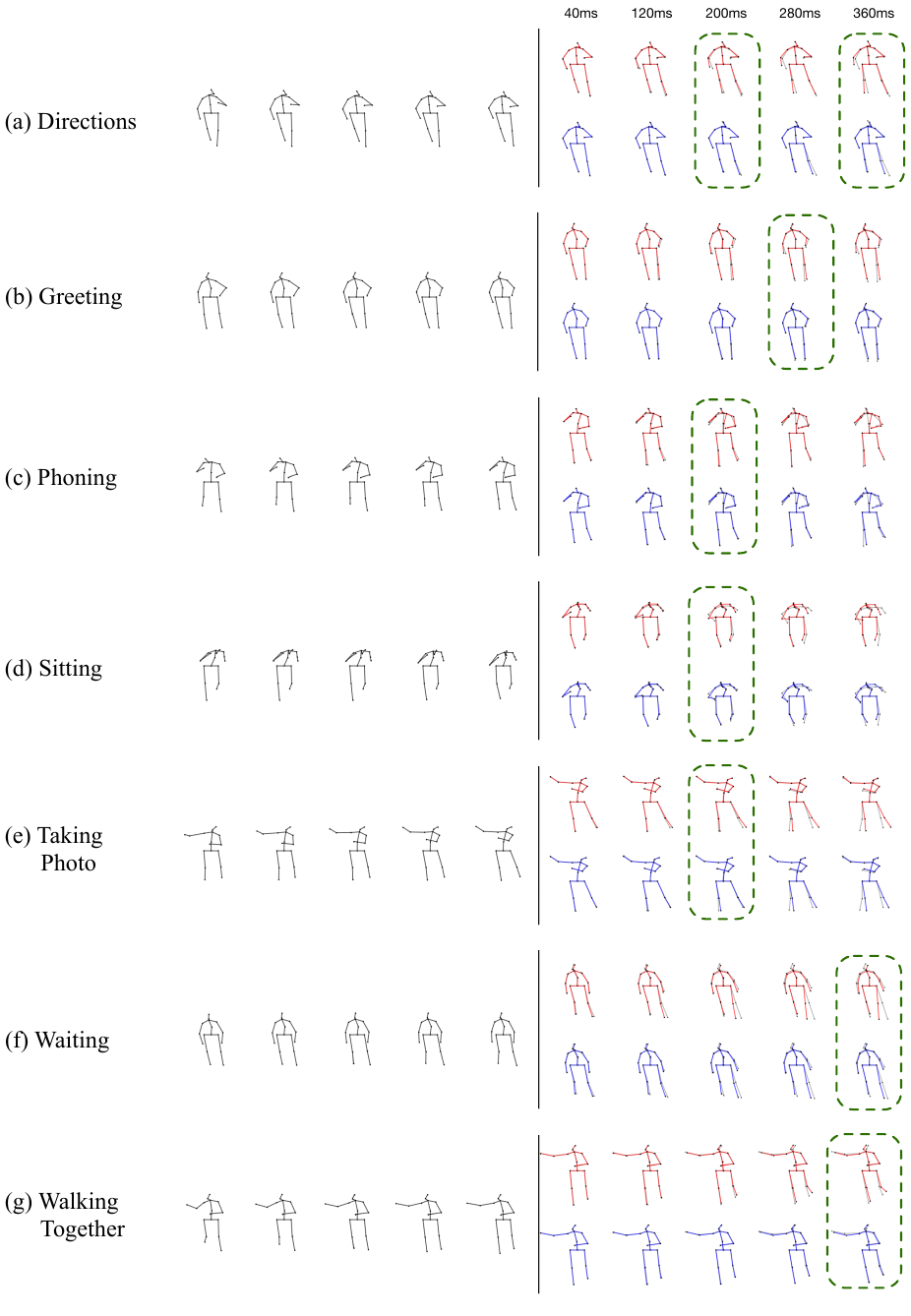}
 \caption{{\bf Qualitative comparison} between (DCT+GCN)\cite{Mao19} (red) and ours (blue) on H3.6M predicting up to $400$ms. The ground truth is superimposed faintly in black on top of both methods. Poses on the left are the conditioning ground truth and the rest are predictions. We observe that our predictions closely match the ground truth poses. We have highlighted some of our best predictions with green bounding boxes.}
\label{fig:qualitative}
 
\end{figure}

We demonstrate our results on the Human 3.6M \cite{Ionescu14a} and CMU Motion Capture\footnote{available at \url{http://mocap.cs.cmu.edu/}} datasets, where we achieve state-of-the-art performance. Qualitative results are shown in Fig.~\ref{fig:qualitative} and \ref{fig:qualitative_lt}. Our code is publicly available at\\ \url{https://github.com/tileb1/motion-prediction-tim}.

\section{Related Work}

The inherent complexity of human motions has driven research towards deep learning models which rely on very large motion capture datasets \cite{Goodfellow16}. Before the deep learning era, analytical models of human motion have been developed by restricting the human motions to simpler or cyclic trajectories like walking \cite{Ormoneit01,Urtasun04a}. However, these models do not generalize well to more complex motions.

\subsubsection{Human motion prediction using RNNs}

Recurrent neural networks (RNN) are standard architectures for modelling time-series data. Since the work of Fragkiadaki \emph{et al.} \cite{Fragkiadaki15}, RNNs have been widely used for human motion forecasting. Jain \emph{et al.} develop S-RNN \cite{Jain16}, which transforms spatio-temporal graphs to a feedforward mixture of RNNs in order to model human motion. Ghosh \emph{et al.} \cite{Ghosh17} propose the Dropout Autoencoder LSTM (DAE-LSTM) to synthesize long-term realistic looking motion sequences.
However the low capacity of the RNN makes it less adequate for high dimensional time-series data like human motion. For instance, Martinez \emph{et al.} \cite{Martinez17b} has shown that RNNs have problems with discontinuity of the predicted sequence at the last seen frame as well as a prediction that converges towards the mean pose of the ground-truth data for long-term predictions. They counter this by adding a residual connection so that the network is made to only predict the residual motion of the subject. More recently, Wang \emph{et al.} \cite{Wang2019ImitationLF} propose a Generative Adversarial Imitation Learning (GAIL) approach for motion prediction. Using GAIL, they iteratively train RNN based policy generator and critic networks. 

\subsubsection{Human motion prediction using other approaches}

There have also been various other architectures proposed for human motion prediction. B{\"u}tepage \emph{et al.} \cite{Butepage17} present several fully-connected encoder-decoder models that aim to encode different properties of the data. One of the models is a time-scale convolutional encoder where they consider different size filters for the input, but not on different length inputs as we propose in our Time Inception Module. In \cite{Butepage18}, conditional variational autoencoder (CVAE) are used to probabilistically model, predict and generate future motions. They extend their probabilistic approach to also incorparate hierarchical action labels in \cite{Butepage19}. Aliakbarian \emph{et al.} \cite{Aliakbarian20} also perform motion generation and prediction by encoding their inputs using a CVAE. They are able to generate diverse motions by randomly sampling and perturbing the conditioning variables.

Closest to our work are Li \emph{et al.} \cite{1805.00655} and Mao \emph{et al.} \cite{Mao19}. Li \emph{et al.} use a convolutional neural network for motion prediction, they produce separate short-term and long-term embeddings. Our Temporal Inception Module also uses convolution operations to produce input embeddings. However our kernel sizes are selected adaptively, and we use the inception module only to capture temporal dependencies within one joint coordinate's trajectory. The dependencies between several trajectories is learned in a separate step through the GCN. Mao \emph{et al.} \cite{Mao19} exploit the graph-like relationship between joints and demonstrate the uses of a GCN for motion prediction. The data undergoes a DCT transformation before being fed to the network, in order to encode the temporal-dependencies within the sequence. Since our embedding strategy also encodes temporal-dependencies, we make use of a similar GCN architecture for the prediction network.

\subsubsection{Inception Module}

The Inception Module was first introduced by Szegedy \emph{et al.} \cite{Szegedy15}, used for the task of object detection and classification. Since then different designs have been proposed \cite{Szegedy16Rethinking} and it has been adapted to a large variety of tasks including human pose estimation \cite{Liu2018ACI}, action recognition \cite{Cho2018SpatioTemporalFN,Hussein19,Yang20}, road segmentation \cite{Doshi2018ResidualIS}, single image super-resolution \cite{Shi2017SingleIS}, and object recognition \cite{Alom2018ImprovedIC}. To the best of our knowledge, we are the first to attempt to modify inception modules for generating input embeddings for motion prediction.

\section{Methodology}
\label{sec:method}

The main encoding methods that have been widely used to represent human motion are 3D joint positions and Euler angle representation. Euler angle representation suffers from ambiguities: two different sets of angles can represent the same pose, which can lead to needlessly over-penalizing predictions. Recent approaches have tried to solve this by changing the encoding to quaternions instead of Euler angles \cite{1805.06485}. For the sake of simplicity, our work is solely based on 3D-joint positions. As such, our data consists of time-sequences of skeletons where each skeleton is encoded as a stack of the 3D encoding of its individual body joints.

Let us now define our task. We are given input sequence of $K$ joint trajectories across time, $\pposes = [\pposes^0, \cdots, \pposes^k, \cdots, \pposes^{K-1}]$, where $k \in \{0, 1, \cdots, {K-1} \}$ represents a Cartesian coordinate value of a joint. Moreover, each joint trajectory $\pposes^{k} = [\pose_{-M}^{k} , \pose_{-M+1}^{k}, \cdots, \pose_{-1}^{k} ]$ is a series of $M$ past joint positions which have already been observed,  where $\pose_i^{k}$ represents a joint coordinate at time index $i$. We aim to predict the poses in the next $T$ frames, $\fposes$. Negative time indices therefore belong to the observed sequence and positive time indices belong to the prediction. For simplicity, we refer to the trajectory of a joint coordinate as "joint trajectory" throughout this paper.

\begin{figure}[t]
  \centering
  \includegraphics[width=\linewidth]{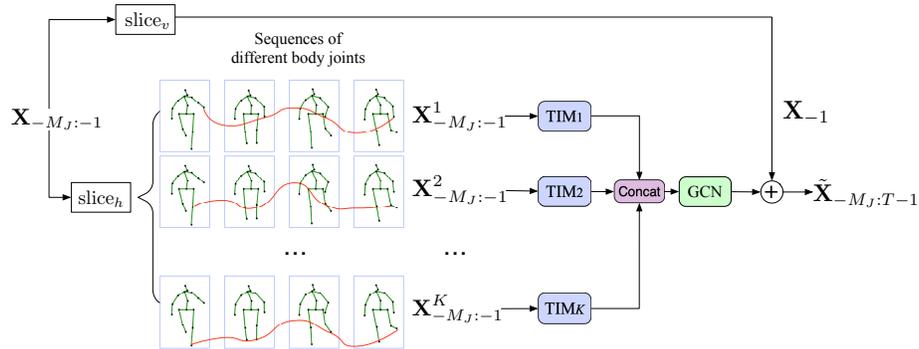}

    \caption{{\bf Overview of the whole framework} making use of multiple TIMs. Using the slice$_h$ operator, we split the input across different joint coordinates. The joint trajectories are fed into the TIMs to produce the embedding, which is then used by the GCN to obtain the residual motion predictions. Using the slice$_v$ operator, separate the most recently seen frame $\textbf{X}_{-1}$, which is broadcasted to all timestamps and summed with the residual GCN results for the final prediction.}
  \label{fig:framework}
\end{figure}

The overall framework converts the input human motion $\pposes$ into embeddings using our temporal inception module (TIM). These embeddings are then fed to the graph convolutional network (GCN) in order to produce the residual motion. The framework is depicted in detail in Fig.~\ref{fig:framework}. The details of the TIM and GCN are introduced below. 

\subsection{Temporal Inception Module}

\begin{figure}[t]
  \centering
  \includegraphics[width=\linewidth]{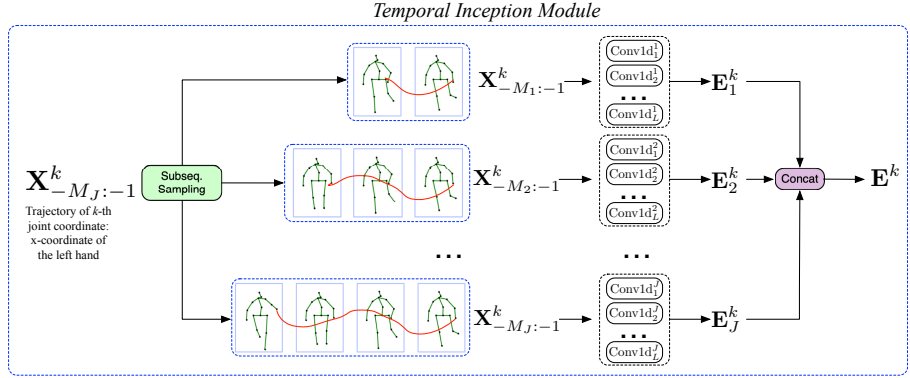}
  \caption{{\bf Overview of the Temporal Inception Module (TIM).} TIM processes each joint coordinate $k$ separately, expressed as a superscript in this figure. The subseq sampling block splits a 1D input sequence into $J$ subsequences, each of length $M_j$. The Conv1d$^j_l$ block corresponds to a 1D convolution operator with kernel size $S^j_l$. The results of the convolutions are concatenated to form the embeddings of each subsequence $\embed_j^k$, which are concatenated again to form the input embeddings  $\embed^k$ to the GCN.}
  \label{fig:time_inception}
\end{figure}

Our main contribution, the Temporal Inception Module (TIM) is illustrated in Fig. \ref{fig:time_inception}. This module is used to obtain embeddings $\embed^k$ of the input motion $\pposes$ for each $k \in \{0, 1, \cdots, {K-1} \}$ joint coordinate.

TIM takes as input a single joint trajectory $\pose^k_{-M_J:-1}$ with the length $M_J$. Then the subsequence sampling block nested in TIM samples the long motion sequence into multiple sequences with different lengths $M_j$ ($M_j > M_i$ if $j > i$).

For example, in our implementation, we consider two different input sizes $M_1=5$ and $M_2=10$ where the past motion the inception module sees are $\pposesA$ and $\pposesB$ respectively. Each input goes through several 1D-convolutions with different sized kernels. The inception module is used to adaptively determine the weights corresponding to these convolution operations. 

Each subsequence $\pose^k_{-M_j:-1}$ has its unique convolutional kernels whose sizes are proportional to the length $M_j$.
In other words, we have smaller kernel size for shorter subsequences and larger kernel size for longer subsequences. The intuition is as follows. 
Using a smaller kernel size allows us to effectively preserve the detailed local information. Meanwhile, for a longer subsequence, a larger kernel is capable of extracting higher-level patterns which depend on multiple time indices. This allows us to process the motion at different temporal scales. 

All convolution outputs are then concatenated into one embedding $\embed^k$ which has the desired features matching our inductive bias i.e. local details for recently seen frames and a low-frequency information for older frames.

More formally, we have
\begin{equation}
    \embed_j^k = \text{concat}(C_{S^j_1}(\pposesj^k), C_{S^j_2}(\pposesj^k), \cdots, C_{S^j_L}(\pposesj^k))
\end{equation}

followed by 

\begin{equation}
    \embed^k = \text{concat}(\embed_1^k, \embed_2^k, \cdots, \embed_J^k)
\end{equation}

where $C_{S^j_l}$ is a 1D-convolution with filter size $S^j_l$. The embedings for each joint trajectory $\embed^k$ are then used as input feature vector for the GCN. An overview of the global framework is illustrated in Fig. \ref{fig:framework}.

\subsection{Graph Convolutional Network}

For the high capacity feed-forward network, we make use a graph convolutional neural network as proposed by Mao et al. \cite{Mao19}. This network is currently a state-of-the-art network for human motion prediction from separate time embeddings of each body joint. This makes it very well suited for our task. As shown in their previous work, using the kinematic tree of the skeleton as predefined weight adgency matrix is not optimal. Instead, a separate adgency matrix is learned for each layer.

Following the notation of \cite{Mao19}, we model the skeleton as a fully connected set of $K$ nodes, represented by the trainable weighted adjacency matrix $\mathbf{A}^{K\times K}$. The GCN consists of several stacked graph convolutional layers, each performing the operation
\begin{align}
    \mathbf{H}^{(p+1)} = \sigma(\mathbf{A}^{(p)}\mathbf{H}^{(p)}\mathbf{W}^{(p)})
\end{align}
where $\mathbf{W}^{(p)}$ is the set of trainable weights of layer $p$, $\mathbf{A}^{(p)}$ is the learnable adgency matrix of layer $p$, $\mathbf{H}^{(p)}$ is the input to layer $p$, $\mathbf{H}^{(p+1)}$ is the output of layer $p$ (and input to layer $p+1$) and $\sigma(\cdot)$ is an activation function. 

The GCN receives as input the embeddings $\embed$ produced by the multiple TIMs and regresses the residual motion which is later summed up with the most recently seen human pose $\pose_{-1}$ to produce the entire motion sequence,

\begin{align}
    \tilde{\mathbf{X}}_{-M_J:T-1} = G(\embed) + \mathbf{X}_{-1}
\end{align}

where the GCN is denoted as $G$. Since $\tilde{\mathbf{X}}_{0:T-1}$ is a subset of $\tilde{\mathbf{X}}_{-M_J:T-1}$, we thus predict the future motion. This is depicted in Fig.~\ref{fig:framework}.

\subsection{Implementation and Training Details}
The Temporal Inception Module used for comparison with other baselines uses 2 input subsequences with lengths $M_1=5$ and $M_2=10$. Both are convolved with different kernels whose sizes are proportional to the subsequence input length. A detailed view of these kernels can be found in Table \ref{tab:tim_detail}. The kernel sizes are indeed chosen to be proportional to the input length. The number of kernels are decreased as the kernel size increases to avoid putting too much weight on older frames. We have also added a special kernel of size $1$ which acts as a pass-through. This leaves us with an embedding $\embed^k$ of size $223$ ($12\cdot4+9\cdot3+9\cdot8+7\cdot6+6\cdot4+1\cdot10$) for each joint coordinate $k \in \{0, 1, \cdots, K-1\}$ which are fed to the GCN. For more details on the GCN architecture, we refer the reader to \cite{Mao19}.

\begin{table}[h!]
\caption{Detailed architecture of Temporal Inception Module used to compare with baselines.}
\label{tab:tim_detail}
\begin{center}
\resizebox{0.7\textwidth}{!}{
\bgroup
\def\arraystretch{1.2}

    \begin{tabular}{C{5cm}|C{3cm}|C{3cm}}
         Subsequence input length ($M_j$) & Number of kernels & Kernel size \\
         \hline
         5 & 12 & 2\\
         5 & 9 & 3\\
         10 & 9 & 3\\
         10 & 7 & 5\\
         10 & 6 & 7\\
         \hline
         10 & 1 & 1\\
    \end{tabular}
\egroup}
\end{center}
\end{table}

The whole network (TIM + GCN) is trained end to end by minimizing the Mean Per Joint Position Error (MPJPE) as proposed in \cite{Ionescu14a}. This loss is defined as

\begin{equation}
    \frac{1}{K(M_J+T)} \sum_{t=-M_J}^{T-1} \sum_{i=1}^{I} ||\mathbf{p}_{i,t}-\mathbf{\hat{p}}_{i,t}||^2
\end{equation}

where $\mathbf{\hat{p}}_{i,t} \in \R^3$ is the prediction of the $i$-th joint at time index $t$, $\mathbf{p}_{i,t}$ is the corresponding ground-truth at the same indices and $I$ is the number of joints in the skeleton ($3\times I=K$ as the skeletons are 3D). Note that the loss sums over negative time indices which belong to the observed sequence as it adds an additional training signal.

It is trained for 50 epochs with a learning-rate decay of 0.96 every 2 epochs as in \cite{Mao19}. One pass takes about 75ms on an NVIDIA Titan X (Pascal) with a batch-size of 16.

\section{Evaluation}
We evaluate our results on two benchmark human motion prediction datasets: Human3.6M \cite{Ionescu14a} and CMU motion capture dataset. The details of the training/testing split of the datasets are shown below, followed by the experimental result analysis and ablation study.

\subsection{Datasets}

\subsubsection{Human3.6M.} Following previous works on motion prediction \cite{Martinez17,Jain16}, we use $15$ actions performed by $7$ subjects for training and testing. These actions are \emph{walking, eating, smoking, discussion, directions, greeting, phoning, posing, purchases, sitting, sitting down, taking photo, waiting, walking dog and walking together}. We also report the average performance across all actions. The 3D human pose is represented using $32$ joints. Similar to previous work, we remove global rotation and translation and testing is performed on the same subset of $8$ sequences belonging to Subject~$5$.

\subsubsection{CMU Motion Capture.} The CMU Motion Capture dataset contains challenging motions performed by $144$ subjects. Following previous related work's training/testing splits and evaluation subset \cite{1805.00655}, we report our results across eight actions: \emph{basketball, basketball signal, directing traffic, jumping, running, soccer, walking, and washwindow}, as well as the average performance. We implement the same preprocessing as the Human3.6M dataset, \emph{i.e.}, removing global rotation and translation. 

\subsubsection{Baselines}

We select the following baselines for comparison: Martinez \emph{et al.} (Residual sup.) in order to compare against the well known method using RNNs \cite{Martinez17b}, Li \emph{et al.} (convSeq2Seq) as they also encode their inputs using convolution operations  \cite{1805.00655} and  Mao \emph{et al.} (DCT+GCN) \cite{Mao19} to demonstrate the gains of using TIM over DCT for encoding inputs. We are unable to compare to the also recent work of \cite{Wang2019ImitationLF} and \cite{Aliakbarian20} due to them reporting results only in joint angle representation and not having code available for motion prediction so far.

\subsection{Results}

In our results (\emph{e.g.}, Tables \ref{tab:ablation_study}, \ref{tab:longterm-h36m}, \ref{tab:shortterm-h36m} and \ref{tab:tim_detail}), for the sake of robustness we report the average error over 5 runs for our own method. We denote our method by ``Ours $(5-10)$'' since our final model takes as input subsequences of lengths $M_1=5$ and $M_2=10$.

We report our short-term prediction results on Human3.6M in Table~\ref{tab:shortterm-h36m}. For the majority of the actions and on average we achieve a lower error than the state-of-the-art (SOTA). Our qualitative results are shown in Figure \ref{fig:qualitative}.

\begin{table}[!ht]
\caption{{\bf Short-term prediction test error of 3D joint positions on H3.6M.} We outperform the baselines on average and for most actions.}
\label{tab:shortterm-h36m}
\begin{center}
\resizebox{1.0\textwidth}{!}{
\bgroup
\def\arraystretch{1.2}

\begin{tabular}{C{3cm}|C{0.9cm}C{0.9cm}C{0.9cm}C{0.9cm}|C{0.9cm}C{0.9cm}C{0.9cm}C{0.9cm}|C{0.9cm}C{0.9cm}C{0.9cm}C{0.9cm}|C{0.9cm}C{0.9cm}C{0.9cm}C{0.9cm}}
\textit{} & \multicolumn{4}{c}{Walking [ms]} & \multicolumn{4}{c}{Eating [ms]} & \multicolumn{4}{c}{Smoking [ms]} & \multicolumn{4}{c}{Discussion [ms]}\\
Name&80&160&320&400&80&160&320&400&80&160&320&400&80&160&320&400\\
\hline
Residual sup. \cite{Martinez17b}&23.8&40.4&62.9&70.9&17.6&34.7&71.9&87.7&19.7&36.6&61.8&73.9&31.7&61.3&96.0&103.5\\
convSeq2Seq \cite{1805.00655}&17.1&31.2&53.8&61.5&13.7&25.9&52.5&63.3&11.1&21.0&33.4&38.3&18.9&39.3&67.7&75.7\\
DCT + GCN \cite{Mao19}&\textbf{8.9}&\textbf{15.7}&\textbf{29.2}&\textbf{33.4}&8.8&18.9&39.4&47.2&7.8&14.9&25.3&\textbf{28.7}&9.8&22.1&\textbf{39.6}&\textbf{44.1}\\
\hline
Ours $(5-10)$&9.3&15.9&30.1&34.1&\textbf{8.4}&\textbf{18.5}&\textbf{38.1}&\textbf{46.6}&\textbf{6.9}&\textbf{13.8}&\textbf{24.6}&29.1&\textbf{8.8}&\textbf{21.3}&40.2&45.5\\
\hline
\end{tabular}
\egroup
 }
\end{center}

\vspace{-0.4cm}

\begin{center}
\resizebox{0.9\textwidth}{!}{
\bgroup
\def\arraystretch{1.2}

\begin{tabular}{C{0.9cm}C{0.9cm}C{0.9cm}C{0.9cm}|C{0.9cm}C{0.9cm}C{0.9cm}C{0.9cm}|C{0.9cm}C{0.9cm}C{0.9cm}C{0.9cm}|C{0.9cm}C{0.9cm}C{0.9cm}C{0.9cm}}
\multicolumn{4}{c}{Directions [ms]} & \multicolumn{4}{c}{Greeting [ms]} & \multicolumn{4}{c}{Phoning [ms]} & \multicolumn{4}{c}{Posing [ms]}\\
80&160&320&400&80&160&320&400&80&160&320&400&80&160&320&400\\
\hline
36.5&56.4&81.5&97.3&37.9&74.1&139.0&158.8&25.6&44.4&74.0&84.2&27.9&54.7&131.3&160.8\\
22.0&37.2&59.6&73.4&24.5&46.2&90.0&103.1&17.2&29.7&53.4&61.3&16.1&35.6&86.2&105.6\\
12.6&24.4&\textbf{48.2}&\textbf{58.4}&14.5&30.5&74.2&89.0&\textbf{11.5}&20.2&\textbf{37.9}&\textbf{43.2}&9.4&23.9&66.2&82.9\\
\hline
\textbf{11.0}&\textbf{22.3}&48.4&59.3&\textbf{13.7}&\textbf{29.1}&\textbf{72.6}&\textbf{88.9}&\textbf{11.5}&\textbf{19.8}&38.5&44.4&\textbf{7.5}&\textbf{22.3}&\textbf{64.8}&\textbf{80.8} \\
\hline
\end{tabular}
\egroup
}
\end{center}

\vspace{-0.4cm}

\begin{center}
\resizebox{0.9\textwidth}{!}{
\bgroup
\def\arraystretch{1.2}

\begin{tabular}{C{0.9cm}C{0.9cm}C{0.9cm}C{0.9cm}|C{0.9cm}C{0.9cm}C{0.9cm}C{0.9cm}|C{0.9cm}C{0.9cm}C{0.9cm}C{0.9cm}|C{0.9cm}C{0.9cm}C{0.9cm}C{0.9cm}}
\multicolumn{4}{c}{Purchases [ms]}&\multicolumn{4}{c}{Sitting [ms]}&\multicolumn{4}{c}{Sitting Down [ms]} & \multicolumn{4}{c}{Taking Photo [ms]}\\
80&160&320&400&80&160&320&400&80&160&320&400&80&160&320&400\\
\hline
40.8&71.8&104.2&109.8&34.5&69.9&126.3&141.6&28.6&55.3&101.6&118.9&23.6&47.4&94.0&112.7\\
29.4&54.9&82.2&93.0&19.8&42.4&77.0&88.4&17.1&34.9&66.3&77.7&14.0&27.2&53.8&66.2\\
19.6&\textbf{38.5}&\textbf{64.4}&\textbf{72.2}&10.7&24.6&50.6&62.0&11.4&\textbf{27.6}&56.4&67.6&6.8&\textbf{15.2}&\textbf{38.2}&\textbf{49.6}\\
\hline
\textbf{19.0}&39.2&65.9&74.6&\textbf{9.3}&\textbf{22.3}&\textbf{45.3}&\textbf{56.0}&\textbf{11.3}&28.0&\textbf{54.8}&\textbf{64.8}&\textbf{6.4}&15.6&41.4&53.5 \\
\hline
\end{tabular}
\egroup
}
\end{center}

\vspace{-0.4cm}

\begin{center}
\resizebox{0.9\textwidth}{!}{
\bgroup
\def\arraystretch{1.2}

\begin{tabular}{C{0.9cm}C{0.9cm}C{0.9cm}C{0.9cm}|C{0.9cm}C{0.9cm}C{0.9cm}C{0.9cm}|C{0.9cm}C{0.9cm}C{0.9cm}C{0.9cm}|C{0.9cm}C{0.9cm}C{0.9cm}C{0.9cm}}
\multicolumn{4}{c}{Waiting [ms]} & \multicolumn{4}{c}{Walking Dog [ms]}&\multicolumn{4}{c}{Walking Together [ms]}&\multicolumn{4}{c}{Average [ms]}\\
80&160&320&400&80&160&320&400&80&160&320&400&80&160&320&400\\
\hline
29.5&60.5&119.9&140.6&60.5&101.9&160.8&188.3&23.5&45.0&71.3&82.8&30.8&57.0&99.8&115.5\\
17.9&36.5&74.9&90.7&40.6&74.7&116.6&138.7&15.0&29.9&54.3&65.8&19.6&37.8&68.1&80.2\\
9.5&22.0&\textbf{57.5}&73.9&32.2&58.0&102.2&122.7&\textbf{8.9}&\textbf{18.4}&\textbf{35.3}&\textbf{44.3}&12.1&25.0&51.0&61.3\\
\hline
\textbf{9.2}&\textbf{21.7}&55.9&\textbf{72.1}&\textbf{29.3}&\textbf{56.4}&\textbf{99.6}&\textbf{119.4}&\textbf{8.9}&18.6&35.5&\textbf{44.3}&\textbf{11.4}&\textbf{24.3}&\textbf{50.4}&\textbf{60.9}\\
\hline
\end{tabular}
\egroup
}
\end{center}

\vspace{0.5cm}

\end{table}

\begin{figure}[!]
  \centering
  \includegraphics[width=1.0\linewidth]{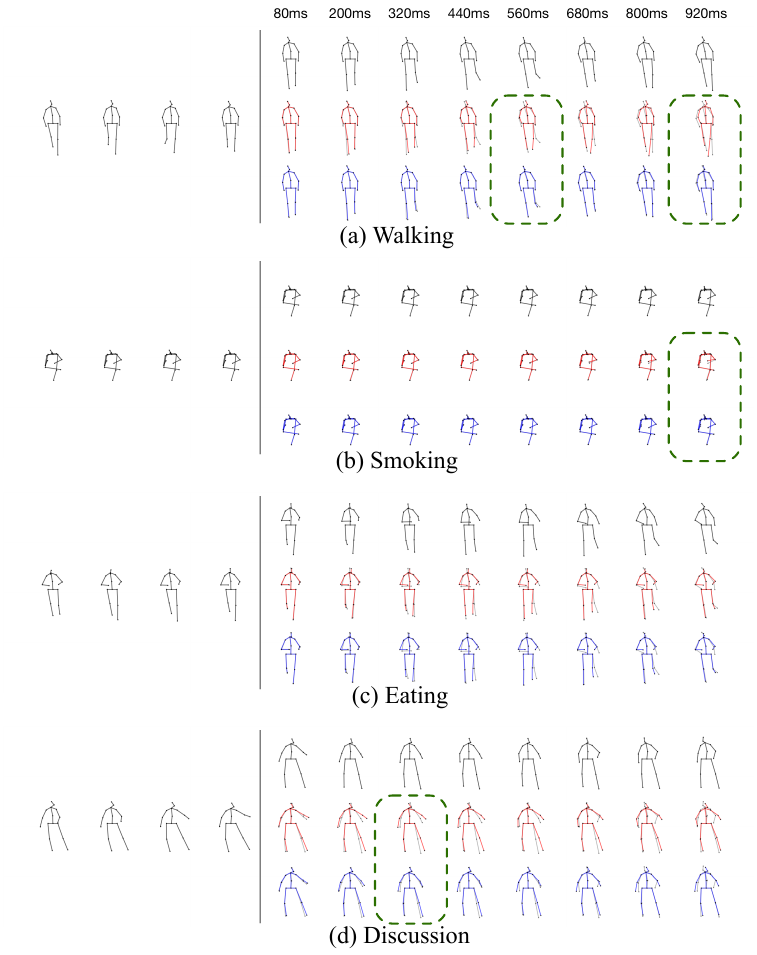}

 \caption{{\bf Long-term qualitative comparison} between ground truth (top row)( DCT+GCN)\cite{Mao19} (middle) and ours (bottom row) on H3.6M predicting up to $1000$ms. The ground truth is superimposed faintly on top of both methods. Poses on the left are the conditioning ground truth and the rest are predictions. We observe that our predictions closely match the ground truth poses, though as expected, the error increases as the time index increases. We have highlighted some of our best predictions with green bounding boxes.}
  \label{fig:qualitative_lt}
\end{figure}

Our long-term predictions on Human3.6M are reported in Table~\ref{tab:longterm-h36m}. Here we achieve an even larger boost in accuracy, especially for case of $1000$ms. We attribute this to the large kernel sizes we have set for input length $10$, which allows the network to pick up the underlying higher-level patterns in the motion. We validate this further in our ablation study. We present our qualitative results in Figure~\ref{fig:qualitative_lt}.

\begin{table}[!]
\caption{{\bf Long-term prediction test error of 3D joint positions on H3.6M.} We outperform the baselines on average and on almost every action. We have also found that we can have an even higher accuracy for $1000$ms in our ablation study, where we show the effect of adding another input subsequence of length $M_j=15$.}
\label{tab:longterm-h36m}
\begin{center}
\resizebox{1.0\textwidth}{!}{
\bgroup
\def\arraystretch{1.2}

\begin{tabular}{C{3cm}|C{1.2cm}C{1.2cm}|C{1.2cm}C{1.2cm}|C{1.2cm}C{1.2cm}|C{1.2cm}C{1.2cm}|C{1.2cm}C{1.2cm}}
\textit{} & \multicolumn{2}{c}{Walking [ms]} & \multicolumn{2}{c}{Eating [ms]} & \multicolumn{2}{c}{Smoking [ms]} & \multicolumn{2}{c}{Discussion [ms]} & \multicolumn{2}{c}{Average [ms]}\\
Name&560&1000&560&1000&560&1000&560&1000&560&1000\\
\hline
Residual sup. \cite{Martinez17b}&73.8 &86.7 &101.3 &119.7 &85.0 &118.5 &120.7 &147.6 &95.2 &118.1\\
convSeq2Seq \cite{1805.00655}&59.2 &71.3 &66.5 &85.4 &42.0 &67.9 &84.1 &116.9 &62.9 &85.4\\
DCT + GCN \cite{Mao19}&42.3 &51.3 &\textbf{56.5} &\textbf{68.6} &\textbf{32.3} &\textbf{60.5} &70.5 &103.5 &50.4 &71.0\\

\hline
Ours $(5-10)$ &\textbf{39.6} &\textbf{46.9} &56.9 &\textbf{68.6} &33.5 &61.7 &\textbf{68.5} &\textbf{97.0} & \textbf{49.6} &\textbf{68.6}\\
\hline
\end{tabular}
\egroup
}
\end{center}
\end{table}

Our predictions on the CMU motion capture dataset are reported in Table~\ref{tab:cmu-acc}. Similar to our results on Human3.6M, we observe that we outperform the state-of-the-art. For all timestamps except for $1000$ ms, we show better performance than the baselines. We observe that both our and Mao \emph{et al.}'s \cite{Mao19} high capacity GCN based models are outperformed by convSeq2Seq \cite{1805.00655}, a CNN based approach. Since the training dataset of CMU-Mocap is much smaller compared to H36M, this leads to overfitting for high-capacity networks such as ours. However, this is not problematic for short-term predictions, as in that case it is not as crucial for the model to be generalizable. We do however outperform Mao \emph{et al.}'s results for the $1000$ms prediction which makes use of the same backbone GCN as us. We observe that on average and for many actions, we outperform the baselines for the $80$, $160$, $320$ and $400$ ms.

\begin{table}[!htbp]
\caption{{\bf Prediction test error of 3D joint positions on CMU-Mocap.} For all timestamps except for $1000$ms, we demonstrate better performance than the baselines. Our model performs better in this case for short term predictions. We observe that on average and for many actions, we surpass the baselines for the $80$, $160$, $320$ and $400$ ms.}   
\label{tab:cmu-acc}
\begin{center}
\resizebox{1.0\textwidth}{!}{
\bgroup
\def\arraystretch{1.2}

\begin{tabular}{C{3cm}|C{0.9cm}C{0.9cm}C{0.9cm}C{0.9cm}C{0.9cm}|C{0.9cm}C{0.9cm}C{0.9cm}C{0.9cm}C{0.9cm}|C{0.9cm}C{0.9cm}C{0.9cm}C{0.9cm}C{0.9cm}}

&\multicolumn{5}{c}{Basketball [ms]} & \multicolumn{5}{c}{Basketball Signal [ms]} & \multicolumn{5}{c}{ Directing Traffic [ms]} \\
Name&80&160&320&400&1000&80&160&320&400&1000&80&160&320&400&1000\\
\hline
Residual sup \cite{Martinez17b}.&18.4&33.8&59.5&70.5&106.7&12.7&23.8&40.3&46.7&77.5&15.2&29.6&55.1&66.1&127.1\\
convSeq2Seq \cite{1805.00655}&16.7&30.5&53.8&64.3&\textbf{91.5}&8.4&16.2&30.8&37.8&76.5&10.6&20.3&38.7&48.4&\textbf{115.5}\\
DCT+GCN \cite{Mao19}&14.0&25.4&49.6&61.4&106.1&3.5&6.1&11.7&\textbf{15.2}&\textbf{53.9}&7.4&15.1&31.7&42.2&152.4\\
\hline
Ours $(5-10)$&\textbf{12.7}&\textbf{22.6}&\textbf{44.6}&\textbf{55.6}&102.0&\textbf{3.0}&\textbf{5.6}&\textbf{11.6}&15.5&57.0&\textbf{7.1}&\textbf{14.1}&\textbf{31.1}&\textbf{41.4}&138.3\\
\hline
\end{tabular}
\egroup
}
\end{center}

\vspace{-0.4cm}

\begin{center}
\resizebox{0.85\textwidth}{!}{
\bgroup
\def\arraystretch{1.2}

\begin{tabular}{C{0.9cm}C{0.9cm}C{0.9cm}C{0.9cm}C{0.9cm}|C{0.9cm}C{0.9cm}C{0.9cm}C{0.9cm}C{0.9cm}|C{0.9cm}C{0.9cm}C{0.9cm}C{0.9cm}C{0.9cm}}

\multicolumn{5}{c}{Jumping [ms]}&\multicolumn{5}{c}{Running [ms]} & \multicolumn{5}{c}{Soccer [ms]}\\
80&160&320&400&1000&80&160&320&400&1000&80&160&320&400&1000\\
\hline
36.0&68.7&125.0&145.5&195.5&15.6&19.4&31.2&36.2&43.3&20.3&39.5&71.3&84&129.6\\
22.4&44.0&87.5&106.3&\textbf{162.6}&\textbf{14.3}&\textbf{16.3}&\textbf{18.0}&\textbf{20.2}&\textbf{27.5}&12.1&21.8&\textbf{41.9}&\textbf{52.9}&\textbf{94.6}\\
16.9&34.4&76.3&96.8&164.6&25.5&36.7&39.3&39.9&58.2&11.3&\textbf{21.5}&44.2&55.8&117.5\\
\hline
\textbf{14.8}&\textbf{31.1}&\textbf{71.2}&\textbf{91.3}&163.5&24.5&37.0&39.9&41.9&62.6&\textbf{11.2}&22.1&45.1&58.1&122.1\\
\hline
\end{tabular}
\egroup
}
\end{center}

\vspace{-0.4cm}

\begin{center}
\resizebox{0.85\textwidth}{!}{
\bgroup
\def\arraystretch{1.2}

\begin{tabular}{C{0.9cm}C{0.9cm}C{0.9cm}C{0.9cm}C{0.9cm}|C{0.9cm}C{0.9cm}C{0.9cm}C{0.9cm}C{0.9cm}|C{0.9cm}C{0.9cm}C{0.9cm}C{0.9cm}C{0.9cm}}

  \multicolumn{5}{c}{Walking [ms]} & \multicolumn{5}{c}{Washwindow [ms]}&\multicolumn{5}{c}{Average [ms]}\\
80&160&320&400&1000&80&160&320&400&1000&80&160&320&400&1000\\

\hline
8.2&13.7&21.9&24.5&\textbf{32.2}&8.4&15.8&\textbf{29.3}&\textbf{35.4}&61.1&16.8&30.5&54.2&63.6&96.6\\
7.6&12.5&23.0&27.5&49.8&8.2&15.9&32.1&39.9&\textbf{58.9}&12.5&22.2&40.7&49.7&\textbf{84.6}\\
7.7&11.8&\textbf{19.4}&23.1&40.2&\textbf{5.9}&\textbf{11.9}&30.3&40.0&79.3&11.5&20.4&37.8&46.8&96.5\\
\hline
\textbf{7.1}&\textbf{11.1}&19.9&\textbf{22.8}&39.3&\textbf{5.9}&12.3&32.1&42.6&80.4&\textbf{10.8}&\textbf{19.5}&\textbf{36.9}&\textbf{46.2}&95.7\\
\hline
\end{tabular}
\egroup
}
\end{center}

\end{table}

\subsection{Ablation Study}

The objective of this section is twofold. 
\begin{itemize}
    \item First, we inquire the effect of choosing a kernel size proportional to the input size $M_j$;
    \item Second, we inquire the effect of the varying length input subsequences.
\end{itemize}

Both results are shown in Fig. \ref{tab:ablation_study}, where the version name represents the set $\{M_j\ : j \in \{1, 2, \cdots, J\}\}$ of varying length subsequences. 

\subsubsection{Proportional filter size}
In our design of TIM, we chose filter sizes proportional to the subsequence input length $M_j$. In Table~\ref{tab:ablation_study}, we observe the effects of setting a ``constant kernel size'' of $2$ and $3$ for all input subsequences. Note that we also adjust the number of filters such that the size of the embedding is the more or less the same for both cases, for fair comparison. We can observe that for both versions $5-10$ and $5-10-15$, having a proportional kernel size to the subsequence input length increases the accuracy for the majority of the actions and this brings better performance on average.  Therefore, our empirical results match our intuition that using larger filters for longer length inputs that look back further into the past helps by capturing higher-level motion patterns which yield embeddings of better quality.

\subsubsection{Varying Length Input Subsequences}
The goal of having the Temporal Inception Module is to sample subsequences of different length $M_j$ which, once processed, yield embeddings with different properties. Embeddings of longer input sequences contain higher level information of the motion (lower frequencies), whereas embeddings of shorter input sequences would contain higher spatial resolution and higher frequency information of the short-term future motion. We expect our model to perform better on very long term prediction of $1000$ms prediction the bigger $M_J$ is. As can be seen from Table \ref{tab:ablation_study}, we also observe that there is unfortunately a trade-off to be made between aiming for very long term predictions ($1000$ms) or shorter term predictions ($560$ms). The $5{-}10{-}15$ model yields higher accuracy than the $5{-}10$ model on $1000$ms and performs worse on $560$ms predictions. This matches our intuition since the $5{-}10{-}15$ model is trained to place more emphasis on the high-level motion pattern and is therefore tuned for very long term predictions at $1000$ms.

Note that we obtain even better performance for very long-term prediction with the $5{-}10{-}15$ model compared with the $5{-}10$ model which has already outperformed the baselines in Table~\ref{tab:longterm-h36m}.

\begin{table}[!htbp]
\caption{{\bf Effect of the kernel size and subsequence lengths $\mathbf{M_j}$} on the framework performance for long-term prediction on H3.6M. We observe that proportional kernel sizes on average yield better performance. We also observe that including the input subsequence with length $M_j=15$ allows us to look back further into the past, boosting the predictions of the furthest timestamp evaluated, $1000$ms.}\label{tab:ablation_study}
\begin{center}
\resizebox{0.95\textwidth}{!}{
\bgroup
\def\arraystretch{1.2}

\begin{tabular}{C{5cm}|C{1.2cm}C{1.2cm}|C{1.2cm}C{1.2cm}|C{1.2cm}C{1.2cm}|C{1.2cm}C{1.2cm}|C{1.2cm}C{1.2cm}}
\textit{} & \multicolumn{2}{c}{Walking [ms]} & \multicolumn{2}{c}{Eating [ms]} & \multicolumn{2}{c}{Smoking [ms]} & \multicolumn{2}{c}{Discussion [ms]} & \multicolumn{2}{c}{Average [ms]}\\
Version&560&1000&560&1000&560&1000&560&1000&560&1000\\
\hline
5-10 (proportional kernel size) &39.6 &46.9 &56.9 &68.6&\textbf{33.5} &\textbf{61.7} &\textbf{68.5} &97.0 & \textbf{49.6} &68.6\\
5-10 (constant kernel size) &\textbf{38.4}&45.6&56.9&68.5&34.9&63.8&73.2&100.1&50.8&69.5\\
\hline
5-10-15 (proportional kernel size) &43.3&43.1&\textbf{45.8}&\textbf{65.2}&36.4&62.9&97.1&\textbf{94.6}&55.7&\textbf{66.5}\\
5-10-15 (constant kernel size) &42.8&\textbf{41.6}&47.1&66.0&36.6&63.2&98.3&96.6&56.2&66.9\\
\end{tabular}
\egroup
}
\end{center}
\end{table}

\section{Conclusion and Future Work}

The task of human motion prediction has gained more attention with the rising popularity of autonomous driving and human-robot interaction. Currently, deep learning methods have made much progress, however, none has focused on utilizing different length input sequences seen at different temporal scales to learn more powerful input embeddings which can benefit the prediction. Our Temporal Inception Module allows us to encode various length input subsequences at different temporal scales and achieves state-of-the-art performance. 

There are many different settings of the Temporal Inception Module to be explored, such as the effects of strided convolutions, allowing for sampling of the input sequence at different rates. The Temporal Inception Module could also be adapted to other applications, such as action recognition. Using longer input subsequences with larger kernels could also be of use for long-term motion generation. We believe these could be interesting avenues for future work and provide further performance gains in their respective fields.

{\small
\bibliographystyle{splncs}
\bibliography{string,vision,graphics,learning,robotics}
}

\end{document}